\documentclass{article} 
\usepackage{iclr2021_conference,times}

\usepackage{amsmath,amsfonts,bm}

\def\eqref#1{equation~\ref{#1}}

\def\1{\bm{1}}

\DeclareMathAlphabet{\mathsfit}{\encodingdefault}{\sfdefault}{m}{sl}
\SetMathAlphabet{\mathsfit}{bold}{\encodingdefault}{\sfdefault}{bx}{n}

\usepackage{hyperref}
\usepackage{url}
\usepackage{graphicx}

\title{Poisoning Deep Reinforcement Learning Agents with In-Distribution Triggers}

\author{Chace Ashcraft, Kiran Karra \\ 
Johns Hopkins University\\
Applied Physics Lab\\
Laurel, MD 20723, USA \\
\texttt{\{chace.ashcraft,kiran.karra\}@jhuapl.edu} \\
}

\iclrfinalcopy 
\begin{document}

\maketitle

\begin{abstract}

In this paper, we propose a new data poisoning attack and apply it to deep reinforcement learning agents. Our attack centers on what we call in-distribution triggers, which are triggers native to the data distributions the model will be trained on and deployed in. We outline a simple procedure for embedding these, and other, triggers in deep reinforcement learning agents following a multi-task learning paradigm, and demonstrate in three common reinforcement learning environments. We believe that this work has important implications for the security of deep learning models.

\end{abstract}

\section{Introduction}

Although deep learning has shown very impressive performance on a plethora of tasks, because of the high cost of data collection, maintenance, and model training, much of the deep learning pipeline is, or could be, outsourced to third parties. This leaves model consumers vulnerable to supply chain vulnerabilities, including backdoor attacks. As part of this work we propose a new attack variant which we believe can be just as detrimental as other attacks while being more difficult to detect or mitigate, even through human inspection. 

Backdoor attacks, also known as trojan attacks, against deep neural networks for classification have been well studied \citep{gu2017badnets}, and many variants of backdoor attacks have been presented \citep{shafahi2018poison, bagdasaryan2020blind, chen2017targeted, kurita2020weight, li2019invisible, liu2020reflection, tran2018spectral, turner2018clean, yao2019latent}. Most of the current research on attacks, detection, and mitigation is focused on classification networks for vision applications.  Some preliminary research into backdoor attacks for text classification networks also exist \citep{dai2019backdoor, sun2020natural}. 


\citet{kiourti2020trojdrl} were among the first to show that DRL agents are also susceptible to poisoning attacks, and outline an algorithm to poison DRL agents under a man-in-the-middle attack model assumption.  In this paper, we expand upon \citet{kiourti2020trojdrl} in the following ways: 1) We demonstrate how to poison deep reinforcement learning (DRL) agents under an alternate attack model, which motivates an important distinction between simple and in-distribution triggers that impacts vulnerability and defense research. 2) We outline and demonstrate a procedure for embedding backdoors into RL agents through the lens of multitask learning, which is interesting due to its simplicity and its connection to another research area in DRL; and provide concrete implementations in popular DRL environments \footnote{Code will be released at \url{https://github.com/trojai}}.


\section{Trojans in Reinforcement Learning Agents}

All methods of embedding backdoors into neural networks involve some manner of data manipulation.  Unlike classification, in DRL, data is generated by an agent interacting with the environment.  TrojDRL~\citep{kiourti2020trojdrl} proposes data poisoning attacks on DRL agents by altering the observations, after they have been generated by the environment, under a man-in-the-middle (MITM) attack model.  More precisely, the observations are altered by a third party, before arriving at the agent to be processed.  The altered observations serve as triggers to the agent to behave with an alternate goal.  

In this work, we consider an attack model where the DRL environments generate poisoned observations. Our contributions are two-fold: 1) Our proposed method of training agents with triggered behavior is simpler than the algorithms outlined by ~\cite{kiourti2020trojdrl}, cast as a multitask learning problem, which is an approach that, to the best of our knowledge, has not been explored by other published methods for this purpose, and 2) It enables further research into triggers which may not be as easily supported by the MITM attack model, such as triggers that may emerge from multiple agents interacting in the environment.


\subsection{Formalism}

The goal of a DRL agent is maximize the total reward received by interacting with its environment, which is typically described as a Markov Decision Process (MDP) ($\mathcal{S}$, $\mathcal{A}$, $\mathcal{P}$, $R$, $\gamma$), where $\mathcal{S}$ is the set of states in the environment, $\mathcal{A}$ is the set of actions (or the action space) available to the agent, $\mathcal{P}: \mathcal{S} \times \mathcal{A} \times \mathcal{S} \longrightarrow [0, 1]$ is the transition probability of going from state $s$ to state $s'$ when taking action $a$, $R: \mathcal{S} \times \mathcal{A} \longrightarrow \mathbb{R}$ is the reward for taking action $a$ from state $s$, and $\gamma \in [0, 1]$ is the discount factor. More accurately, many DRL environments are better represented as Partially Observable Markov Decision Process (POMDP), ($\mathcal{S}$, $\mathcal{A}$, $\mathcal{P}$, $R$, $\Omega$, $O$, $\gamma$), where $\Omega$ is a set of possible observations that depend on the state of the environment, and $O: \mathcal{O} \times \mathcal{A} \times \mathcal{S} \longrightarrow [0, 1]$ is the probability of receiving observation $o$ given that the agent is in state $s'$ after taking action $a$. 

\subsection{Triggers}

Behavior change in poisoned RL agents is enabled by a trigger in the agent's observation.  A trigger is defined as an attribute of the agent's observation that causes the agent to behave with an alternative goal.  Triggers can occur in whatever representation the RL agent uses to accomplish its goal; the two representations that we consider in this paper are state-space and visual representations.  State-space representations are vectors which contain meaningful information about the state of the environment and are usually expertly defined, whereas visual representations are images of the agent's field of view.

For these two representations, we differentiate between simple triggers and in-distribution triggers.  A simple trigger refers to any change to an observation which can cause the agent to behave differently.  Examples of simple triggers in state-space may include adding or multiplying a constant to the state vector, and applying the modulo operator to be within the agent's input space.  In the vision space, simple triggers may include changes to inconspicuous pixels, such as artificially altering the color of the top right pixel in the environment.  While these triggers highlight the vulnerability of DRL agents to small perturbations in their observations, they require unnatural modifications to the RL environment.

In contrast, we define an in-distribution trigger as changes to an observation that are not anomalous to the environment.  More formally, borrowing from POMDP terminology, let $\Omega$ to be the set of possible observations that the environment can produce.  We define in-distribution triggers to be patterns in observations contained in $\Omega$, and simple triggers to be patterns in observations not contained in $\Omega$.

Examples of in-distribution triggers in the state-space are specific patterns of input which can occur as a result of agent interaction with the environment, and which may have a specific semantic meaning.  For example, suppose the state-space representation for an agent playing Atari Breakout is a vector where the first element represents the ball speed, the second element represents the ball direction, the third represents the ball location, and the fourth represents the agent location.  An in-distribution trigger, here, could be if the ball location, speed, and direction are equal to a specific, preprogrammed value.  Similarly, an example of the in-distribution trigger in the image-space representation could be a certain configuration of the color of game objects, if that information is in the agent's field of view.  Note that in-distribution triggers do not require modifications to the RL environment observations, although RL environments could be designed to contain usable in-distribution triggers.  We hypothesize that this makes in-distributions more difficult to detect, because trigger detection methods typically rely on some manner of anomaly detection \citep{xu2019detecting, chen2018detecting, chen2019deepinspect, huang2019neuroninspect, liu2019abs}, and consequently more dangerous.


\subsection{Training Methodology}

As a proof of concept, we embedded both simple and in-distribution triggers in DRL agents using a training strategy akin to multitask learning \citep{mnih2015human} with Proximal Policy Optimization (PPO) \citep{schulman2017proximal}  \footnote{We used an implementation of PPO from the \textbf{torch\_ac} module publicly available (\url{{https://github.com/lcswillems/torch-ac}}) for all of our examples.}. If we consider the clean behavior as one task, and the trojan, or poisoned, behavior as another, then the problem of embedding the trigger reduces to a multitask learning problem, which we solve by training on both triggered environments and clean environments in parallel. We found this strategy to work very reasonably, when the number of triggered and clean environments was tuned to ensure that both tasks were learned. We refer to this as \textit{data balancing.}

Inspired by previous work with embedding trojans in classification models \citep{karra2020trojai}, we attempt to balance the amount of clean and triggered data that the RL agent receives during training. We found that having an appropriate balance was often critical for efficient learning. For example, 10\%-20\% triggered data often lead to successful training, which implemented by training on 2 triggered environments and 8 clean environments in parallel. 

\subsection{Examples}

\subsubsection{Simple Trigger: Atari Boxing}

We implemented two simple triggers via a modified version of OpenAI Gym's \citep{openai_gym} version of Atari Boxing. The first uses the game's RAM observations (i.e. ``Boxing-ram-v0''), and the trigger is to set the observation to $10 \times$ the original observation modulo 255. The second trigger uses RGB images (i.e. ``Boxing-v0''), with the trigger being the top left corner greyed out, similar to \cite{kiourti2020trojdrl}. We produce trojaned behavior by negating the reward when the trigger is active (i.e. the environment is triggered or poisoned). 

We use a fully-connected network for the RAM observations, and a CNN for the RGB observations. The fully connected network consists of a shared embedding that is two hidden layers of 512 nodes with ReLU activations, and an actor and critic each consisting of one hidden layer of 512 nodes and ReLU activations. The CNN is similar, but has three convolution layers with ReLU activations as the shared embedding. Observations are resized to 48x48 gray-scale images before being passed into the network. 

Training required 21 million frames of data for the RAM model and 25 million frames for the image model. We used a split of 8 clean, 2 triggered environments for each observation type. The resulting performances where scores of 52 average reward for clean environments for both types of observations, and -66 and -60 average reward for triggered environments for RAM and image observation types, respectively.

\subsubsection{In-Distribution Trigger: Parameterized LavaWorld}

In order to demonstrate an example of the in-distribution trigger, we introduce the Parameterized LavaWorld environment, which is a modified version of the \texttt{MiniGrid-LavaCrossingS9N1-v0} environment from \texttt{gym-minigrid}~\citep{gym_minigrid} Python module. We modify the original environment to include an extra line of 3 lava squares, and parameterize the placement of the main ``river'' of lava, the location of the gap in the river, and the location of the extra lava squares, via arguments to the reset method. 

\begin{figure}
    \centering
    \includegraphics[width=0.55\linewidth]{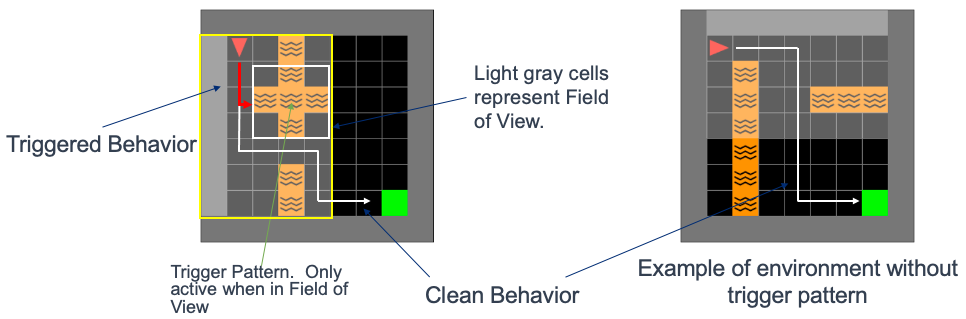}
    \caption{The goal of Parameterized Lavaworld is to get to the green goal square without stepping in an orange lava square. The trigger is a cross or ``T'' pattern (left) of lava squares, such that when the agent sees the pattern, it enters one of the pattern's squares instead of going to the goal.}
    \label{fig:param_lw}
\end{figure}

The trigger occurs when the extra line of lava squares forms a cross or ``T'' with the main river; Figure~\ref{fig:param_lw}. If the agent sees this cross in one of its observations, the reward function returns a positive reward only if the agent enters one of the lava squares in the trigger. Otherwise, the reward function remains the same. By definition, this trigger is in-distribution because it is a valid observation of the normal environment. Indeed, the same environment can be run without the reward modification, and would be similar to the original \texttt{gym-minigrid} LavaCrossing tasks. 

This environment may output a simplified, but image like, (7, 7, 3) state observation, or an RGB image observation. We were able to embed a trigger for both observation types, but focus here on simplified observation case. We trained a standard actor-critic model with a CNN based image embedding but omitting the recurrent network \footnote{ \url{https://github.com/lcswillems/rl-starter-files/blob/master/model.py}}, using 10 environments in parallel where the trigger was active in each, and trained for 40 million frames. The resulting agent successfully navigated to the goal in 97 of the 100 test episodes when the trigger was not present, and navigated to the trigger pattern in 94 of the 100 test episodes when the trigger was observed. Further, we believe we could increase both clean and triggered performance via basic hyperparameter tuning. 


\subsection{In-Distribution Trigger: Pursuit}

For a deep multi-agent reinforcement learning (DMARL) example, we consider the Pursuit environment from the PettingZoo library \citep{terry2020pettingzoo}, which consists of various ``pursuer'' agents that must work together in a 2D space in order to surround other ``evader'' agents. The pursuers are controlled by DMARL agents, and the evaders move randomly. 

To insert a trigger we first follow the approach in \cite{terry2020parameter} to train clean agents using the deep reinforcement learning package RLlib \citep{liang2018rllib}. For simplicity, we configure the environment to have 4 agents.  During training, we only populate the environment with 1 evader and we set all ``touch'' rewards, which are given when a pursuer is next to an evader, to 0. We found this helped force the agents to learn to seek out evaders and surround them.

We then insert a trigger by fine-tuning a trained, clean model for a single pursuer using an altered reward function. The triggered reward structure provides positive reward (the same as the surround reward) when the agent sees itself one step away from surrounding the evader, but then receives a negative reward for the actual surround, and is 0 otherwise. We train the triggered agent for approximately 1.2 million frames, after which we observe mean triggered reward converge to around 250, meaning that the triggered agent is frequently entering the ``almost surround'' state without making the surrounds. 


\section{Conclusion}

In this paper, we have introduced an alternate backdoor attack using in-distribution triggers and a new, but simple, training method for embedding backdoors into RL agents. We hypothesize that in-distribution triggers, due to their nature, are more difficult to detect and present a new problem for the security community to address. The problem seems particularly significant given the ease with which these triggers may be embedded. 

\subsubsection*{Acknowledgments}
This research is based upon work supported in part by the Office of the Director of National Intelligence (ODNI), Intelligence Advanced Research Projects Activity (IARPA).

\bibliography{iclr2021_conference}
\bibliographystyle{iclr2021_conference}



\end{document}